%% file: main.tex
\documentclass{article}

\usepackage[nonatbib,final]{cls/neurips_workshop_2021}

\usepackage[utf8]{inputenc} 
\usepackage[T1]{fontenc}    
\usepackage{hyperref}       
\usepackage{url}            
\usepackage{booktabs}       
\usepackage{amsfonts}       
\usepackage{nicefrac}       
\usepackage{microtype}      
\usepackage{xcolor}         

\usepackage{times}

\usepackage{enumitem}
\usepackage[noadjust]{cite}
\usepackage{amsmath,amssymb,amsfonts,amsthm,amscd,dsfont} 
\usepackage{algorithm,algorithmicx,listings}        
\usepackage[noend]{algpseudocode}			        

\usepackage{graphicx,tabularx,adjustbox}
\usepackage{multicol}
\usepackage[font={small}]{caption}   
\usepackage{subcaption}

\newtheorem*{proposition*}{Proposition}

\newtheorem*{corollary*}{Corollary}

\theoremstyle{definition}

\newtheorem*{assumption*}{Assumption}
\newtheorem*{problem*}{Problem}
\newtheorem{problem}{Problem}
\theoremstyle{remark}

\newtheorem*{solution*}{Solution}

\newcommand{\prl}[1]{\left(#1\right)}

\DeclareMathOperator*{\tr}{tr}
\DeclareMathOperator*{\diag}{diag}
\newcommand{\scaleMathLine}[2][1]{\resizebox{#1\linewidth}{!}{$\displaystyle{#2}$}}

\newcommand{\NEW}[1]{{\color{black}#1}}

\input{cls/sym.tex}

\title{Physics-guided \NEW{Learning-based} Adaptive Control on the SE(3) Manifold}

\author{Thai Duong \qquad\qquad   Nikolay Atanasov\\
	Department of Electrical and Computer Engineering\\
	University of California San Diego \\
	La Jolla, CA 92093, USA \\
	e-mail: {\tt\small \{tduong,natanasov\}@ucsd.edu}
}

\begin{document}
	
	\maketitle
	
	\begin{abstract}
		\NEW{In real-world robotics applications, accurate models of robot dynamics are critical for safe and stable control in rapidly changing operational conditions. This motivates the use of machine learning techniques to approximate robot dynamics and their disturbances over a training set of state-control trajectories. This paper demonstrates that inductive biases arising from physics laws can be used to improve the data efficiency and accuracy of the approximated dynamics model. For example, the dynamics of many robots, including ground, aerial, and underwater vehicles, are described using their $SE(3)$ pose and satisfy conservation of energy principles. We design a physically plausible model of the robot dynamics by imposing the structure of Hamilton's equations of motion in the design of a neural ordinary differential equation (ODE) network. The Hamiltonian structure guarantees satisfaction of $SE(3)$ kinematic constraints and energy conservation by construction. It also allows us to derive an energy-based adaptive controller that achieves trajectory tracking while compensating for disturbances. Our learning-based adaptive controller is verified on an under-actuated quadrotor robot.}
	\end{abstract}
	

\input{tex/Introduction.tex}

\input{tex/ProblemStatement.tex}
	\input{tex/TechnicalApproach.tex}
\input{tex/Experiments.tex}
	\input{tex/Conclusion.tex}

	%

	\bibliographystyle{cls/IEEEtran}
	\bibliography{bib/thai_ref.bib}

\end{document}

%% file: cls/sym.tex
\newcommand{\calD}{{\cal D}}

\newcommand{\calH}{{\cal H}}

\newcommand{\calL}{{\cal L}}

\newcommand{\frakp}{{\mathfrak{p}}}
\newcommand{\frakq}{{\mathfrak{q}}}

\newcommand{\bfa}{\mathbf{a}}

\newcommand{\bfd}{\mathbf{d}}
\newcommand{\bfe}{\mathbf{e}}
\newcommand{\bff}{\mathbf{f}}
\newcommand{\bfg}{\mathbf{g}}

\newcommand{\bfp}{\mathbf{p}}

\newcommand{\bfr}{\mathbf{r}}

\newcommand{\bfu}{\mathbf{u}}
\newcommand{\bfv}{\mathbf{v}}

\newcommand{\bfx}{\mathbf{x}}

\newcommand{\bfzeta}{{\boldsymbol{\zeta}}}

\newcommand{\bftheta}{{\boldsymbol{\theta}}}

\newcommand{\bfpi}{{\boldsymbol{\pi}}}
\newcommand{\bfrho}{{\boldsymbol{\rho}}}

\newcommand{\bftau}{{\boldsymbol{\tau}}}

\newcommand{\bfphi}{{\boldsymbol{\phi}}}

\newcommand{\bfpsi}{{\boldsymbol{\psi}}}
\newcommand{\bfomega}{{\boldsymbol{\omega}}}

\newcommand{\bfI}{\mathbf{I}}
\newcommand{\bfJ}{\mathbf{J}}
\newcommand{\bfK}{\mathbf{K}}

\newcommand{\bfM}{\mathbf{M}}

\newcommand{\bfR}{\mathbf{R}}

\newcommand{\bfV}{\mathbf{V}}
\newcommand{\bfW}{\mathbf{W}}

\newcommand{\bbR}{\mathbb{R}}

%% file: tex/Introduction.tex
\section{Introduction}
\label{sec:intro}

Autonomous mobile robots operating in \NEW{real-world} complex and dynamic conditions require accurate dynamics models \cite{ljung1999system, loquercio2021autotune} for motion planning and control. This has motivated the development of data-driven approaches to learn dynamics models \cite{deisenroth2015gp,berkenkamp16safe,chang2017jump,hewing2019cautious,torrente2021data, raissi2018multistep,chua2018deep, nguyen2011model} and disturbance models \cite{sanner1991gaussian, gahlawat20al1adaptive, grande2013nonparametric, joshi2019deep,joshi2020asynchronous, o2021meta, harrison2018control, richards21adaptive} from data. As data-driven dynamics models require large amounts of data, inductive biases such as physical knowledge about the system has been imposed on the model structure \cite{sanchez2018graph, greydanus2019hamiltonian, lutter2019deepunderactuated, zhong2020symplectic, duong21hamiltonian}, which a black-box model might struggle to infer. Such structures also simplify the design of a nominal stable regulation or tracking controller \cite{lutter2019deepunderactuated, zhong2020symplectic, duong21hamiltonian}. 



This paper develops \NEW{physics-guided} data-driven adaptive control for rigid-body systems, such as unmanned ground vehicle (UGVs), unmanned aerial vehicle (UAVs), or unmanned underwater vehicles (UUVs), that satisfy Hamilton's equations of motion over the $SE(3)$ manifold. \NEW{Given a dataset of state-control trajectories with different disturbance realizations, we impose Hamilton's equations of motion on the architecture of a neural ODE network to \emph{learn physically plausible system dynamics and disturbance model}. The learned system dynamics respect kinematic and energy conservation constraints by construction and formulated in terms of \emph{interpretable quantities}: mass and inertia matrix, potential energy, input gain matrix, and disturbance features. \emph{The interpretability of the model enables us to design an adaptive controller}, consisting of an energy-based tracking controller for the learned model and an adaptation law that compensates disturbances online by scaling the learned disturbance features using the $SE(3)$ tracking error. We verify our dynamics learning and control approach on an under-actuated quadrotor robot in the PyBullet simulator \cite{gym-pybullet-drones2020}.}

%% file: tex/ProblemStatement.tex
\section{Problem Statement}
\label{sec:problem_statement}
Consider a robot modeled as a single rigid body with position $\bfp \in \bbR^3$, orientation $\bfR\in SO(3)$, body-frame linear velocity $\bfv \in \bbR^3$, and body-frame angular velocity $\bfomega \in \bbR^3$. Let $\frakq = [\bfp^\top\;\; \bfr_1^\top\;\; \bfr_2^\top\;\; \bfr_3^\top]^\top \in \bbR^{12}$ be the generalized coordinates, where $\bfr_1$, $\bfr_2$, $\bfr_3$ are the rows of the rotation matrix $\bfR$. Let $\bfzeta = [\bfv^\top\;\;\bfomega^\top]^\top \in \bbR^6$ be the generalized velocity. The state $\bfx = (\frakq, \bfzeta)$ evolves on the tangent bundle $TSE(3)$ of the pose manifold $SE(3)$ and is governed by the robot dynamics: $\dot{\bfx} = \bff(\bfx, \bfu, \bfd)$, where $\bfu$ is the control input and $\bfd$ is a disturbance signal. The disturbance $\bfd$ appears as an external force applied to the system and is modeled as a linear combination of nonlinear features $\bfW(\frakq, \frakp)\in \bbR^{6\times p}$: $\bfd(t) = \bfW(\frakq(t), \frakp(t))\bfa^*$
%
%
where $\bfa^* \in \bbR^p$ are unknown weights. 

As a mechanical system, the robot obeys Hamilton's equations of motion \cite{lee2017global}. The Hamiltonian, $\mathcal{H}(\mathbf\frakq, \mathbf\frakp) = \frac{1}{2}\mathbf\frakp^\top \bfM(\mathbf\frakq)^{-1} \mathbf\frakp + V(\mathbf\frakq)$, captures the total energy of the system as the sum of the kinetic energy $T(\mathbf\frakq, \mathbf\frakp) =  \frac{1}{2}\mathbf\frakp^\top \bfM(\mathbf\frakq)^{-1} \mathbf\frakp$ and the potential energy $V(\frakq)$, where $\frakp = \bfM(\frakq)\bfzeta \in \bbR^6$ is the generalized momentum and $\bfM(\frakq) \in \bbR^{6\times6}$ is the generalized mass matrix.
%
%
%
%
%
The system dynamics are governed by Hamilton's equations:
\begin{equation}
\label{eq:PH_dyn}
\begin{bmatrix}
\dot{\mathbf\frakq} \\
\dot{\mathbf\frakp} \\
\end{bmatrix}
= \begin{bmatrix}
\bf0 & \mathbf\frakq^{\times} \\
-\mathbf\frakq^{\times\top} & \mathbf\frakp^{\times} 
\end{bmatrix}
\begin{bmatrix}
\frac{\partial \mathcal{H}}{\partial\mathbf\frakq} \\
\frac{\partial \mathcal{H}}{\partial\mathbf\frakp} 
\end{bmatrix} + \begin{bmatrix} \bf0 \\ \bfg(\mathbf\frakq) \end{bmatrix}\bfu + \begin{bmatrix} \bf0 \\ \bfd \end{bmatrix}.
\end{equation}
The cross maps $\mathbf\frakq^{\times}$ and $\mathbf\frakp^{\times}$ are defined as $\mathbf\frakq^{\times} = \begin{bmatrix}
\bfR^\top\!\!\!\! & \bf0 & \bf0 & \bf0 \\
\bf0 & \hat{\bfr}_1^\top & \hat{\bfr}_2^\top & \hat{\bfr}_3^\top
\end{bmatrix}^\top\!\!\!\!, \quad \mathbf\frakp^{\times} = \begin{bmatrix} \mathbf\frakp_{\bfv}\\\mathbf\frakp_{\bfomega}\end{bmatrix}^{\times} \!\!\!\!= \begin{bmatrix}
\bf0 & \hat{\mathbf\frakp}_{\bfv}\\
\hat{\mathbf\frakp}_{\bfv} & \hat{\mathbf\frakp}_{\bfomega}
\end{bmatrix},$
where the hat map $\hat{(\cdot)}: \bbR^3 \mapsto \mathfrak{so}(3)$ constructs a skew-symmetric matrix from a 3D vector. The time derivative of the generalized velocity is:
\begin{equation}
\label{eq:PH_zetadot}
\dot{\bfzeta} =  \prl{ \frac{d}{dt} \bfM^{-1}(\mathbf\frakq) }\mathbf\frakp + \bfM^{-1}(\mathbf\frakq)\dot{\mathbf\frakp}.
\end{equation}
In summary, \eqref{eq:PH_dyn} and \eqref{eq:PH_zetadot} capture the structure of the system dynamics $\dot{\bfx} = \bff(\bfx,\bfu,\bfd)$.


Consider a collection $\calD = \{\calD_1, \calD_2, \ldots, \calD_M\}$ of state-control trajectory datasets $\calD_j$, each collected under a different unknown disturbance realization $\bfa_j^*$, for $j = 1,\ldots,M$. A trajectory dataset $\calD_j = \{t^{(ij)}_{0:N}, \bfx^{(ij)}_{0:N}, \bfu^{(ij)}\}_{i = 1}^{D_j}$ consists of $D_j$ state sequences $\bfx^{(ij)}_{0:N}$, obtained by applying a constant control input $\bfu^{(ij)}$ to the system with initial condition $\bfx^{(ij)}_0$ at time $t_0^{(ij)}$ and sampling its state $\bfx^{(ij)}(t_n^{(ij)}) =: \bfx^{(ij)}_n$ at times $t_{0:N}^{(ij)}$. Our objective is to approximate the dynamics $\bff$ by $\bar{\bff}_{\bftheta}$, where the parameters $\bftheta$ characterize the unknowns $\bfM_\bftheta(\frakq)$, $V_\bftheta(\frakq)$, $\bfg_\bftheta(\frakq)$, as well as the disturbance model by $\bfW_\bfphi(\frakq, \frakp) \bfa_j$, where the parameters $\bfphi$, $\{\bfa_j\}_{j=1}^M$ model each disturbance sample. To optimize $\bftheta$, $\bfphi$, $\{\bfa_j\}$, we predict the state sequence $\bar{\bfx}^{(ij)}_{1:N}$ using the approximated dynamics starting from state $\bfx^{(ij)}_0$ with a constant control $\bfu^{(ij)}$ and minimize the loss as defined in Problem \ref{problem:dynamics_learning}.


\begin{problem}
	\label{problem:dynamics_learning}
	Given $\calD = \{\{t^{(ij)}_{0:N}, \bfx^{(ij)}_{0:N}, \bfu^{(ij)}\}_{i = 1}^{D_j}\}_{j=1}^M$, find parameters $\bftheta$, $\bfphi$, $\{\bfa_j\}_{j=1}^M$ that minimize:
	\begin{align} \label{problem_formulation_unknown_env_equation}
	\min_{\bftheta, \bfphi, \{\bfa_j\}} \;&\sum_{j=1}^M \sum_{i = 1}^{D_j} \sum_{n = 1}^N \ell(\bfx^{(ij)}_n,\bar{\bfx}^{(ij)}_n)\notag\\
	\text{s.t.} \;\; & \dot{\bar{\bfx}}^{(ij)}(t) = \bar{\bff}_{\bftheta}(\bar{\bfx}^{(ij)}(t), \bfu^{(ij)}, \bar{\bfd}^{(ij)}(t)),\;\;\bar{\bfd}^{(ij)}(t) = \bfW_\bfphi(\bar{\bfx}^{(ij)}(t)) \bfa_j,\\
	&\bar{\bfx}^{(ij)}(t_0) = \bfx^{(ij)}_0,\;\;\bar{\bfx}^{(ij)}_n = \bar{\bfx}^{(ij)}(t_n), \forall j = 1, \ldots, M, \;\;\forall n = 1, \ldots, N,\;\;\forall i = 1, \ldots, D_j,\notag
	\end{align}
	where $\ell$ is a distance metric on the state space $TSE(3)$.
\end{problem}

After the offline system identification in Problem~\ref{problem:dynamics_learning}, we aim to design a controller $\bfu = \bfpi(\bfx, \bfx^*, \bfa; \bftheta, \bfphi)$ that tracks a desired state trajectory $\bfx^*(t)$, $t \geq t_0$, using the learned dynamics $\bar{\bff}_{\bftheta}$ and disturbance $\bfW_\bfphi$ models. To handle disturbance realizations $\bfd = \bfW_\bfphi(\bfx)\bfa^*$ with unknown ground-truth $\bfa^*$, we augment the tracking controller with an adaptation law $\dot{\bfa} = \bfrho(\bfx, \bfx^*, \bfa; \bfphi)$ that estimates $\bfa^*$ online.

%% file: tex/TechnicalApproach.tex
\section{Technical Approach}
\label{sec:technical_approach}
We present our approach in two stages: Hamiltonian-based dynamics and disturbance model learning (Problem \ref{problem:dynamics_learning}) and adaptive control design with disturbance compensation.

\begin{figure*}[!t]
	\centering
	\includegraphics[width=0.5\textwidth]{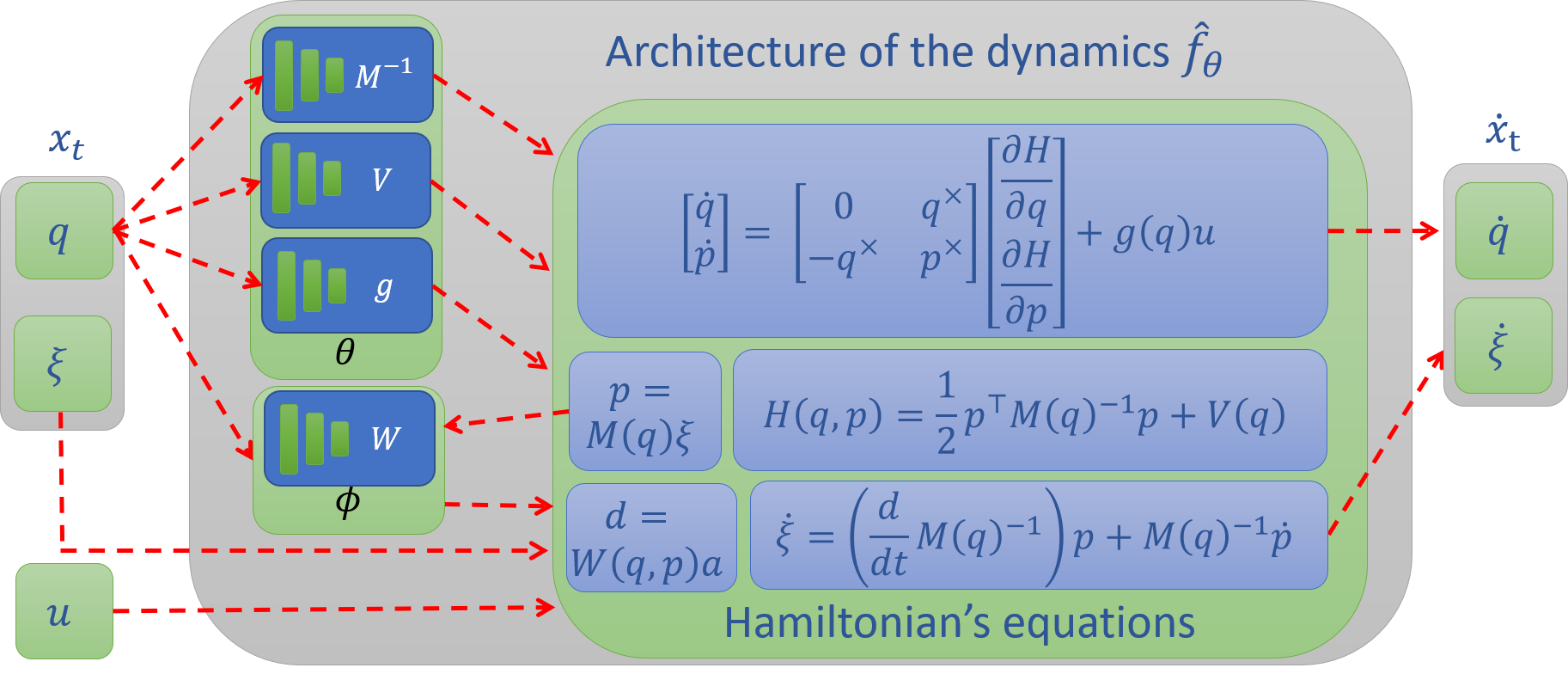}%
	\caption{Architecture of the proposed physics-guided neural ODE networks.}
	\label{fig:arch}
\end{figure*}

\textbf{Hamiltonian-based dynamics learning with disturbances.}
We use a neural ODE network \cite{chen2018neural} whose structure (Fig. \ref{fig:arch}) respects Hamilton's equations \eqref{eq:PH_dyn}\eqref{eq:PH_zetadot}  and, hence, guarantees satisfaction of the $SE(3)$ kinematic constraints and energy conservation by construction. Specifically, the unknowns $\bfM_\bftheta(\frakq)$, $\bfV_\bftheta(\frakq)$ and $\bfg_\bftheta(\frakq)$ are approximated by three separate neural networks. We extend the work on Hamiltonian-based neural ODE learning \cite{duong21hamiltonian} by introducing a disturbance model, $\bfd = \bfW_\bfphi(\frakq, \frakp) \bfa$, described by another neural network, and estimating its parameters $\bfphi$, $\bfa$ along with the system dynamics parameters $\bftheta$.
We first define the $TSE(3)$ distance metric $\ell$ in Problem~\ref{problem:dynamics_learning} as $\ell(\bfx,\bar{\bfx}) = \ell_{\bfp}(\bfx,\bar{\bfx}) + \ell_{\bfR}(\bfx,\bar{\bfx}) + \ell_{\bfzeta}(\bfx,\bar{\bfx}),$
%
%
%
where the position, orientation, and velocity errors are defined respectively as $\ell_{\bfp}(\bfx,\bar{\bfx}) = \| \bfp - \bar{\bfp}\|^2_2, \ell_{\bfR}(\bfx,\bar{\bfx}) = \|\left(\log (\bar{\bfR} \bfR^\top)\right)^{\vee} \|_2^2, \ell_{\bfzeta}(\bfx,\bar{\bfx}) = \| \bfzeta - \bar{\bfzeta}\|^2_2$. 
%
%
The $\log$-map $\log : SE(3) \mapsto \mathfrak{so}(3)$ is the inverse of the exponential map, associating a rotation matrix to a skew-symmetric matrix and $(\cdot)^\vee : \mathfrak{so}(3) \mapsto \bbR^3$ is the inverse of the hat map $\hat{(\cdot)}$.
Let $\calL(\bftheta, \bfphi, \{\bfa_j\}; \calD)$ be the total loss in Problem~\ref{problem:dynamics_learning}. To calculate the loss, for each dataset $\calD_j$ with disturbance $\bar{\bfd}^{(ij)}(t) = \bfW_\bfphi(\bar{\bfx}^{(ij)}(t)) \bfa_j$, we solve an ODE: $\dot{\bar{\bfx}}^{(ij)} = \bar{\bff}_{\bftheta}(\bar{\bfx}^{(ij)}, \bfu^{(ij)}, \bar{\bfd}^{(ij)}), \quad \bar{\bfx}^{(ij)}(0) = \bfx^{(ij)}_{0}$, 
%
%
using an ODE solver \cite{chen2018neural}. This generates a predicted system state trajectory $\bar{\bfx}^{(ij)}_{1:N}$ at times $t^{(ij)}_{1:N}$ for each $i = 1,\ldots, D_j$ and $j = 1, \ldots, M$, sufficient to compute $\calL(\bftheta, \bfphi, \{\bfa_j\}; \calD)$. The parameters $\bftheta$, $\bfphi$, and $\bfa_j$ are updated using gradient descent by back-propagating the loss through the neural ODE solver.

\textbf{Physics-guided data-driven adaptive control.}
Enabled by the learned Hamiltonian-based dynamics, we develop a tracking controller $\bfpi$ using the interconnection and damping assignment passivity-based control (IDA-PBC) method \cite{van2014port}. Consider a desired state trajectory $\bfx^*(t) = (\mathbf\frakq^*(t), \bfzeta^*(t))$. The desired momentum is calculated as  $\mathbf\frakp^* = \bfM (\frakq) \begin{bmatrix} \bfR^\top \bfR^* \bfv^* \\ \bfR^\top \bfR^* \bfomega^* \end{bmatrix}$ by transforming the desired velocity $\bfzeta^* = [\bfv^{*\top}\;\; \bfomega^{*\top}]^\top$ to the body frame.
%
%
As the Hamiltonian of the system is not necessarily minimized along $(\mathbf\frakq^*(t), \mathbf\frakp^*(t))$, the key idea of the IDA-PBC design is to choose the control input $\bfu(t)$ so that the closed-loop system has a desired Hamiltonian $\calH_d(\mathbf\frakq, \mathbf\frakp)$ that is minimized along $(\mathbf\frakq^*(t), \mathbf\frakp^*(t))$. Using a quadratic error on the tangent bundle $TSE(3)$, we design the desired Hamiltonian:
\begin{equation} \label{eq:desired_hamiltonian}
\mathcal{H}_d(\mathbf\frakq, \mathbf\frakp) =  \frac{1}{2}k_\bfp(\bfp - \bfp^*)^\top(\bfp - \bfp^*) + \frac{1}{2} k_{\bfR}\tr(\bfI - \bfR^{*\top}\bfR) + \frac{1}{2}(\mathbf\frakp-\mathbf\frakp^*)^\top\bfM^{-1}(\mathbf\frakq)(\mathbf\frakp-\mathbf\frakp^*),
\end{equation}
where $k_\bfp$ and $k_{\bfR}$ are positive gains. 
The controller $\bfpi(\bfx,\bfx^*,\bfa ; \bftheta, \bfphi)$ consists of an energy-shaping term $\bfu_{ES}$, a damping-injection term $\bfu_{DI}$, and a disturbance compensation term $\bfu_{DC}$:
\begin{equation}
\label{eq:ES_DI_COMP_control}
\begin{aligned}
\bfu_{ES} &=\scaleMathLine[0.5]{\bfg^{\dagger}(\mathbf\frakq)\left(\mathbf\frakq^{\times\top} \frac{\partial V}{\partial \mathbf\frakq} - \mathbf\frakp^{\times}\bfM^{-1}(\mathbf\frakq)\mathbf\frakp - \bfe(\mathbf\frakq,\mathbf\frakq^*) + \dot{\frakp}^*\right)},\\
\bfu_{DI} &= -\bfK_\bfd\bfg^{\dagger}(\mathbf\frakq)   \bfM^{-1}(\mathbf\frakq)(\mathbf\frakp-\mathbf\frakp^*), \text{and } \bfu_{DC}= -\bfg^{\dagger}(\mathbf\frakq)\bfW(\mathbf\frakq, \mathbf\frakp) \bfa
\end{aligned}
\end{equation}
where $\bfg^{\dagger}(\mathbf\frakq)$ is the pseudo-inverse of $\bfg(\mathbf\frakq)$, $\bfK_\bfd = \diag(k_\bfv \bfI, k_\bfomega\bfI)$ is a damping gain with positive terms $k_\bfv$, $k_\bfomega$, the coordinate error is  $\bfe(\mathbf\frakq,\mathbf\frakq^*) = \scaleMathLine[0.45]{\begin{bmatrix} k_\bfp \bfe_{\bfp}(\mathbf\frakq,\mathbf\frakq^*) \\ k_{\bfR}\bfe_{\bfR}(\mathbf\frakq,\mathbf\frakq^*) \end{bmatrix} = \begin{bmatrix} k_\bfp \bfR^\top(\bfp - \bfp^*) \\ \frac{1}{2}k_{\bfR}\prl{\bfR^{*\top}\bfR-\bfR^\top\bfR^{*}}^{\vee}\end{bmatrix}}$,
%
%
and the  momentum error is $ \mathbf\frakp - \mathbf\frakp^* =  \scaleMathLine[0.45]{\bfM(\mathbf\frakq) \begin{bmatrix} \bfe_{\bfv}(\bfx,\bfx^*) \\ \bfe_{\bfomega}(\bfx,\bfx^*) \end{bmatrix}= \bfM(\mathbf\frakq) \begin{bmatrix} \bfv - \bfR^\top \bfR^* \bfv^* \\ \bfomega - \bfR^\top \bfR^* \bfomega^* \end{bmatrix}}.$
%
%
%
Please refer to \cite{duong21hamiltonian} for a detailed derivation of $\bfu_{ES}$ and $\bfu_{DI}$.

The disturbance compensation term $\bfu_{DC}$ in \eqref{eq:ES_DI_COMP_control} requires online estimation of the disturbance feature weights $\bfa$. Inspired by \cite{slotine1989composite}, we design an adaptation law which utilizes the body-frame geometric errors  $\bfe_\bfp, \bfe_\bfR, \bfe_\bfv,$ and $\bfe_\bfomega$ to update the disturbance feature weights:
\begin{equation}
\label{eq:geometric_adaptive_law}
\dot{\bfa} = \bfrho(\bfx,\bfx^*,\bfa; \bfphi)  = \bfW_{\bfphi}^\top(\mathbf\frakq, \mathbf\frakp) \begin{bmatrix} c_{\bfp} \bfe_{\bfp}(\mathbf\frakq,\mathbf\frakq^*) + c_{\bfv} \bfe_{\bfv}(\bfx,\bfx^*) \\ c_{\bfR} \bfe_{\bfR}(\mathbf\frakq,\mathbf\frakq^*) + c_{\bfomega} \bfe_{\bfomega}(\bfx,\bfx^*) \end{bmatrix},
\end{equation}
where $c_{\bfp}$, $c_{\bfv}$, $c_{\bfR}$, $c_{\bfomega}$ are positive coefficients.

%% file: tex/Experiments.tex
\begin{figure*}[!t]
	\centering

	\begin{minipage}[c]{0.8\textwidth}
		\begin{subfigure}[t]{0.5\textwidth}
		\centering
		\includegraphics[width=\textwidth]{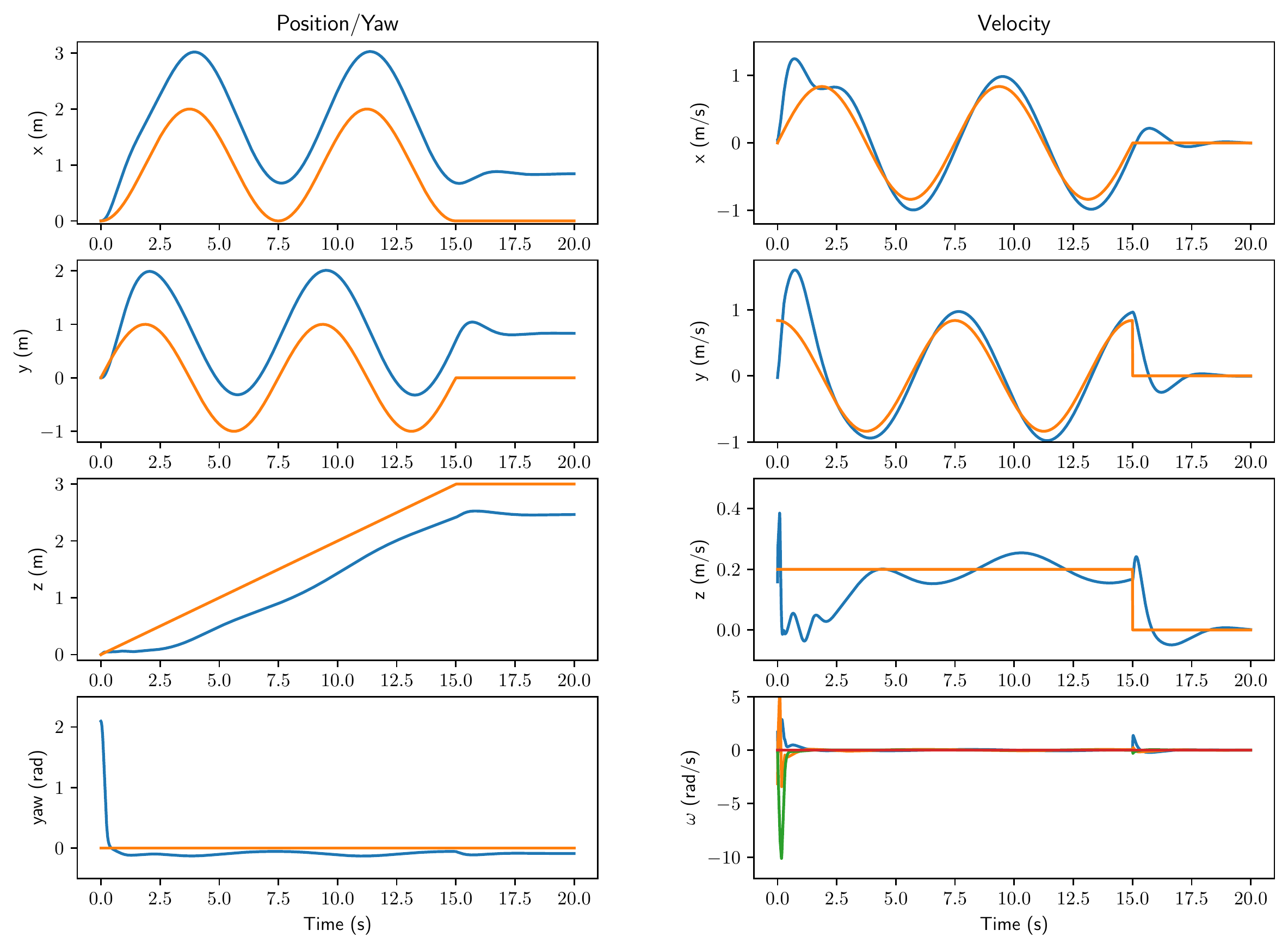}%
		\caption{}
		\label{fig:tracking_results_noadaptive_exp1}
		\end{subfigure}%
	\hfill
		\begin{subfigure}[t]{0.5\textwidth}
		\centering
		\includegraphics[width=\textwidth]{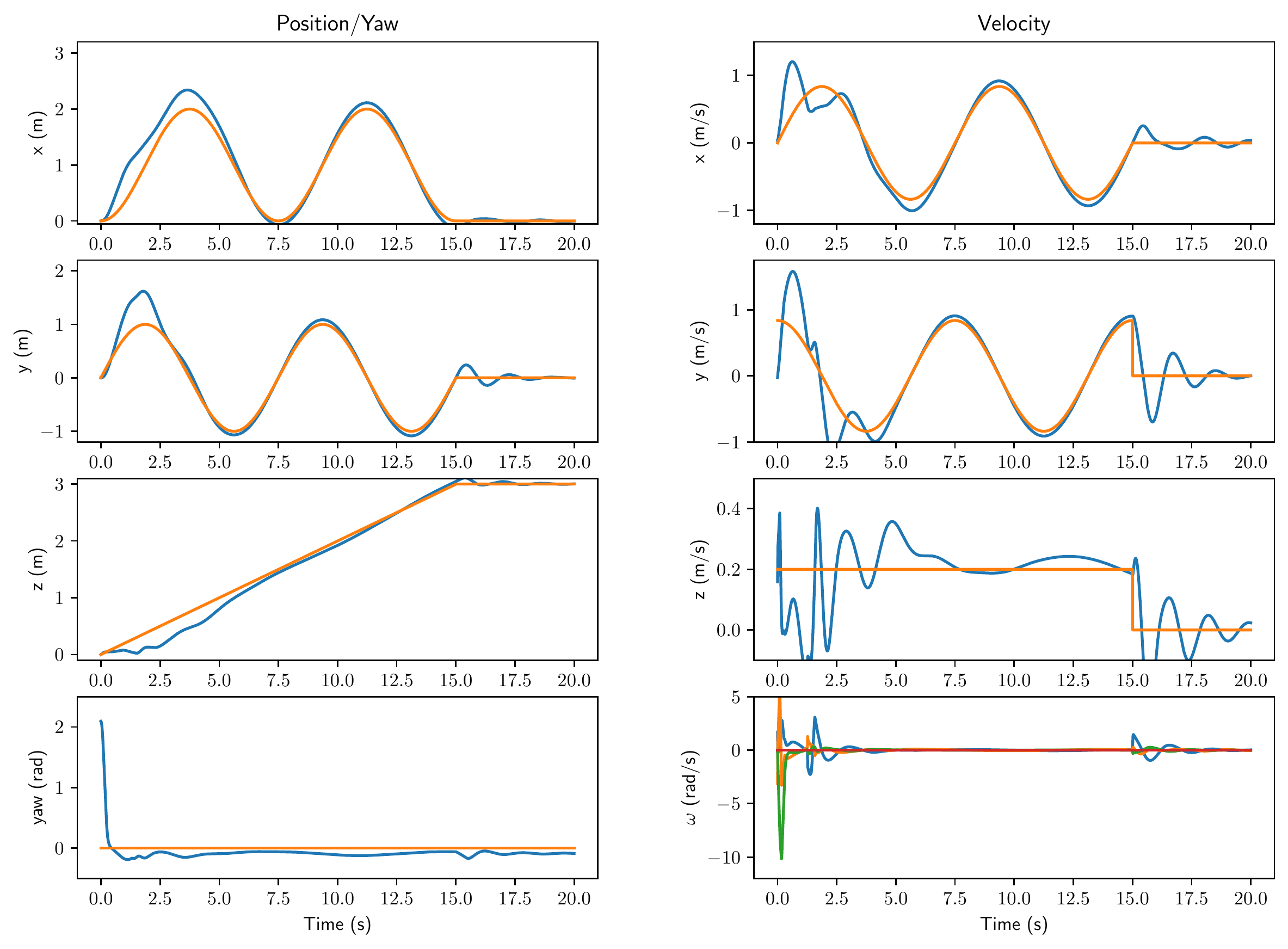}%
		\caption{}
		\label{fig:tracking_results_adaptive_exp1}
	\end{subfigure}%
	\end{minipage}
	\begin{minipage}[c]{0.19\textwidth}
		\begin{subfigure}[t]{\textwidth}
			\centering
			\includegraphics[width=0.85\textwidth]{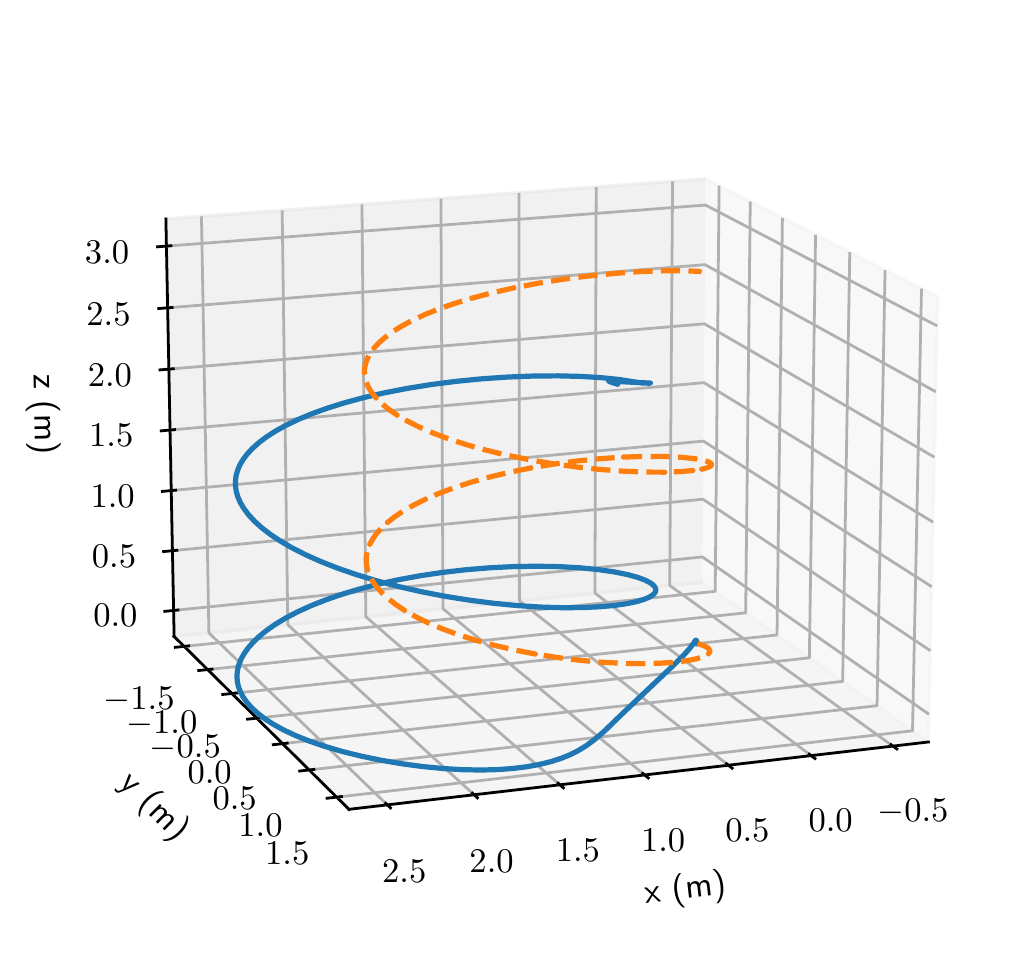}\\
			\includegraphics[width=0.85\textwidth]{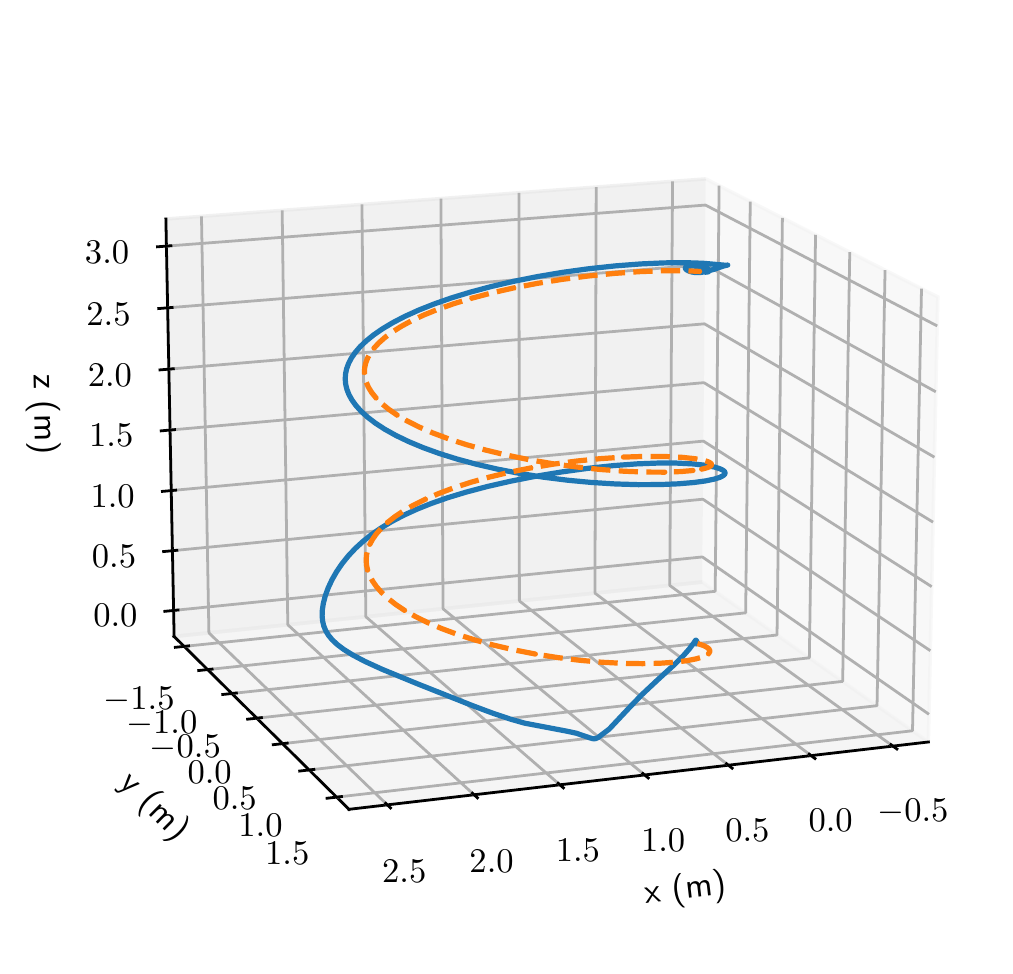}%
			\caption{}
			\label{fig:pybullet_trajviz}
		\end{subfigure}%
	\end{minipage}
	\caption{Tracking a spiral trajectory with a PyBullet Crazyflie quadrotor \cite{gym-pybullet-drones2020} with disturbances from constant wind and defective rotors: (a)(c)-upper without adaptation and (b)(c)-lower with adaptation. With adaptation, our controller is able to estimate and compensate for the disturbances and track the trajectory successfully.}
	\label{fig:quadrotor_exp}
\end{figure*}

\section{Evaluation}
\label{sec:exp_results}

We consider a simulated PyBullet Crazyflie quadrotor \cite{gym-pybullet-drones2020}, with control input $\bfu = [f, \bftau]$ including a thrust $f\in \mathbb{R}_{\geq 0}$ and a torque vector $\bftau \in \mathbb{R}^3$ generated by the $4$ rotors.  The disturbance $\bfd$ comes from two sources: 1) horizontal winds, simulated as an external force $\bfd_w = \begin{bmatrix}
d_{wx} & d_{wy} & 0
\end{bmatrix}^\top \in \bbR^3$ in the world frame, i.e. $\bfR^\top\bfd_w$ in the body frame, applied on the quadrotor, and 2) two defective rotors $1$ and $2$, generating $\delta_1$ and $\delta_2$ percents of the nominal thrust, respectively. We collect a dataset $\calD = \{\calD_j\}_{j = 1}^M$ with $M = 8$ realizations of the disturbance $\bfd_{wj}$, $\delta_{1j}$ and $\delta_{2j}$. Specifically, the wind components $w_{xj}, w_{yj}$ are chosen from the set $\{\pm 0.25, \pm 0.5\}$ while the values of $\delta_{1j}$ and $\delta_{2j}$ are sampled from the range $[94\%, 98\%]$. For each disturbance realization, a PID controller  provided by \cite{gym-pybullet-drones2020} is used to drive the quadrotor from a random starting point to $9$ different desired poses, generating a dataset $\mathcal{D}_j = \{t_{0:N}^{(ij)},\mathbf\frakq_{0:N}^{(ij)}, \bfzeta_{0:N}^{(ij)}, \bfu^{(ij)})\}_{i=1}^{D_j}$ with $N = 5$ and $D_j = 1080$.

Since the mass $m$ of the quadrotor is easily measured, we use the ground-truth value $m = 0.027$kg and form our approximated generalized mass matrix as $\bfM_\bftheta(\mathbf\frakq) =\diag(m\bfI, \bfJ_{\bftheta}(\mathbf\frakq))$,
where $\bfJ_{\bftheta}(\mathbf\frakq)$ represents an unknown inertial matrix. The known mass $m$ leads to a known potential energy $V(\frakq) = mg\begin{bmatrix}
0 & 0 & 1
\end{bmatrix} \bfp$, where $\bfp$ is the position of the quadrotor and $g\approx 9.8ms^{-2}$ is the gravitational acceleration. As described in Sec. \ref{sec:technical_approach}, we learn $\bfM_\bftheta(\mathbf\frakq)$, $\bfg_\bftheta(\frakq)$ and $\bfW_\bfphi(\frakq, \frakp)$ from the dataset $\calD$.

We verify our learned adaptive controller in Sec. \ref{sec:technical_approach} by driving the quadrotor to track a pre-defined trajectory in the present of the aforementioned disturbance $\bfd$. The desired trajectory is specified by the desired position $\bfp^*(t)$ and the desired heading $\bfpsi^*(t)$, which is used to construct an appropriate choice of $\bfR^*$ and $\bfomega^*$, similar to \cite{lee2010geometric,goodarzi2015geometric,duong21hamiltonian}. The tracking controller \eqref{eq:ES_DI_COMP_control}, with gains $k_\bfp = 0.135, k_\bfv = 0.0675, k_{\bfR} = 1.0,$ and $ k_{\bfomega} = 0.08$ is paired with the adaptation law \eqref{eq:geometric_adaptive_law} with gains $c_{\bfp}~=~0.04, c_{\bfv} = 0.04$, and $c_{\bfR} = c_{\bfomega} = 0.5$.
We test the learned adaptive controller with a task of tracking a spiral trajectory with a constant wind $\bfd_w = \begin{bmatrix}
0.75 & 0.75 & 0
\end{bmatrix}$, and defective rotors $1\&2$ with $\delta_1 = \delta_2 = 0.8$. The quadrotor without adaptation drifts as seen in Fig. \ref{fig:tracking_results_noadaptive_exp1} and \ref{fig:pybullet_trajviz} (upper). Meanwhile, Fig. \ref{fig:tracking_results_adaptive_exp1} and \ref{fig:pybullet_trajviz} (lower) show that our adaptive controller is able to estimate the disturbance online after a few seconds and successfully tracks the trajectory.

%% file: tex/Conclusion.tex
\section{Conclusion}
\label{sec:conclusion}

This paper introduced a physics-guided gray-box model for rigid-body dynamics learning from disturbance-corrupted data. We developed a Hamiltonian neural ODE architecture which captures external force disturbances and respects the $SE(3)$ constraints and energy conservation by construction. We designed an energy-based tracking controller and an adaptation law that compensates for disturbances relying on geometric tracking errors, and verified their effectiveness on a quadrotor.